\def\year{2019}\relax
\documentclass[letterpaper]{article} %
\usepackage{aaai19}  %
\usepackage{times}  %
\usepackage{helvet} %
\usepackage{courier}  %
\usepackage[hyphens]{url}  %
\usepackage{graphicx} %
\usepackage{subfig}
\usepackage{graphicx}  %
\usepackage{times}

\usepackage{soul}
\usepackage{url}
\usepackage[utf8]{inputenc}
\usepackage{graphicx}
\usepackage{amsmath}
\usepackage{booktabs}
\usepackage{algorithm}
\usepackage{algorithmic}
\usepackage{xspace} 
\usepackage{amssymb}
\usepackage{bm}
\usepackage{color}
\usepackage{url}
\usepackage{booktabs}

\title{Terminal Prediction as an Auxiliary Task for Deep Reinforcement Learning}
\author{Bilal Kartal,\thanks{Equal contribution} Pablo Hernandez-Leal$^*$ and Matthew E. Taylor\\
Borealis AI\\
Edmonoton, Canada\\
\{bilal.kartal, pablo.hernandez, matthew.taylor\}@borealisai.com
}

\begin{document}

\maketitle

\begin{abstract}

Deep reinforcement learning has achieved great successes in recent years, but there are still open challenges, such as convergence to locally optimal policies and sample inefficiency. In this paper, we contribute a novel self-supervised auxiliary task, i.e., \emph{Terminal Prediction} (TP), estimating temporal closeness to terminal states for episodic tasks. The intuition is to help representation learning by letting the agent predict how close it is to a terminal state, while learning its control policy. Although TP could be integrated with multiple algorithms, this paper focuses on Asynchronous Advantage Actor-Critic (A3C) and demonstrating the advantages of A3C-TP. Our extensive evaluation includes: a set of Atari games, the BipedalWalker domain, and a mini version of the recently proposed multi-agent Pommerman game. Our results on Atari games and the BipedalWalker domain suggest that A3C-TP outperforms standard A3C in most of the tested domains and in others it has similar performance. In Pommerman, our proposed method provides significant improvement both in learning efficiency and converging to better policies against different opponents.
\end{abstract}

\section{Introduction}

\noindent Deep reinforcement learning (DRL) combines reinforcement learning~\cite{sutton2018reinforcement} with deep learning~\cite{lecun2015deep}, enabling better scalability and generalization for high-dimensional domains. DRL has been one of the most active areas of research in recent years with great successes such as mastering Atari games from only raw images~\cite{mnih2015human}, a Go playing agent skilled well beyond any human player~\cite{silver2017mastering}, and very recently, great successes in multiagent games (e.g., DOTA 2 and Starcraft II). Reinforcement learning has also been applied for interactive narrative generation~\cite{wang2017interactive} and learning companion NPC (Non-Player Character) behaviors~\cite{sharifi2010learning}.  For a more general and technical overview of DRL, please see the recent comprehensive surveys~\cite{arulkumaran2017deep,franccois2018introduction,hernandez2018multiagent}.

One of the biggest challenges for reinforcement learning is sample efficiency~\cite{yu2018towards}. Data hungriness of model-free RL methods is only aggravated when the reward signals are sparse, delayed, or noisy. Standard RL problem formulations with non-linear function approximation (i.e., DRL) combine representation learning together with policy learning. In this case, the problem is further deepened when rewards are sparse since most of the collected experiences do not produce a learning signal for the agent, thus delaying representation learning for the environment. To address these issues, the concept of \emph{auxiliary tasks} was introduced where an RL agent can learn from all experiences independent of external reward signals~\cite{shelhamer2016loss,sutton2018reinforcement}. \emph{Auxiliary tasks} can be any task that the RL agent can predict and observe from the environment in a self-supervised fashion such as reward prediction~\cite{jaderberg2016reinforcement}, or predicting a game specific feature such as presence/absence of enemies in the current observation~\cite{lample2017playing}. Also note that \emph{auxiliary tasks} are different from model-based RL where the learned model is used for planning~\cite{oh2015action,leibfried2016deep}. In contrast, auxiliary tasks were originally presented as \emph{hints} that improved the network performance and learning time~\cite{suddarth1990rule}. In a minimal example of a small neural network it was shown that adding an auxiliary task effectively removed local minima. Thus, the auxiliary losses are expected to give more ambient gradients, not necessarily yield a generative model~\cite{shelhamer2016loss}. 

Different auxiliary tasks have been successfully evaluated such as: reward prediction~\cite{jaderberg2016reinforcement}, modeling the inverse dynamics of the environment~\cite{shelhamer2016loss}, and depth prediction~\cite{mirowski2016learning}. In contrast, we propose \emph{Terminal Prediction (TP)}, where the intuition is to help the representation learning by letting the agent predict how close it is to a terminal state while learning the standard actor policy. TP targets are similarly computed in a self-supervised fashion, but they are independent of reward sparsity of the game or any other domain dynamics that might render representation learning challenging such as drastic changes in domain visuals. 

In this paper, we consider Asynchronous Advantage Actor-Critic (A3C)~\cite{mnih2016asynchronous} as a baseline algorithm as it is one of the frontier approaches among asynchronous distributed deep RL techniques. Then, we make the following contributions:

\begin{itemize}

\item We propose a novel auxiliary task, namely \emph{Terminal Prediction (TP)}, aiming to enhance the RL agent with a capability of predicting a measure of temporal closeness to expected terminal states, i.e. likely to be reached with the current agent policy, without extra signals from the environment. In this work, we propose A3C-TP, which results from integrating TP task to A3C with minimal refinements to ensure that it is still on-policy. Note that even though we showcase TP with A3C, it can be combined with other deep RL methods. 

\item We conduct experiments on a diverse set of Atari games and the BipedalWalker domain where A3C-TP either outperforms or performs similar to the standard A3C method. We also conduct experiments on a mini version of a recently released multi-agent domain, i.e. Pommerman~\cite{resnick2018pommerman}, showing that A3C-TP both learns faster and converges to better policies, compared to A3C, against different opponents.
\end{itemize}

\section{Related Work}

Reinforcement learning approaches mainly fall under three categories: value-based methods such as Q-learning~\cite{watkins1992q} or Deep-Q Network~\cite{mnih2015human}; policy-based methods such as REINFORCE~\cite{williams1992simple}; and a combination of value- and policy-based techniques, i.e. actor-critic methods~\cite{konda2000actor}. In particular, in the last category several distributed actor-critic based DRL algorithms have been recently proposed~\cite{jaderberg2016reinforcement}. One notable example is A3C (Asynchronous Advantage Actor-Critic)~\cite{mnih2016asynchronous}, which is an algorithm that employs a \emph{parallelized} asynchronous training scheme (using multiple CPU cores) for efficiency.

Recently, auxiliary tasks have been proposed to improve representation learning in deep RL. For example, Mirowski et al.~(\citeyear{mirowski2016learning}) studied self-supervised tasks in a navigation problem. Their results show that augmenting the RL agent with auxiliary tasks supports representation learning which provides richer training signals that enhance data efficiency. Lample and Chaplot~(\citeyear{lample2017playing}) proposed to add an auxiliary task to a DRL agent in the Doom game; in particular, the agent was trained to predict the presence/absence of enemies in the current observation. Lastly, a concurrent work~\cite{leibfried2018modelbased} proposed a model based DRL architecture based on Deep-Q-Network which predicts: Q-values, next frame, rewards, and a binary terminal flag that predicts whether the episode will end or not. The terminal flag is similar to our \emph{Terminal Prediction} except that (i) our method is not model-based, and (ii) we formulate the prediction problem as a regression problem rather than classification, thus we gather a lot more self-supervised signals that are automatically class-balanced (this is particularly essential for tasks with long episodes).

Another related work to ours is the UNREAL framework~\cite{jaderberg2016reinforcement} which is built on top of the A3C with several refinements and \emph{auxiliary task} integration. In particular, UNREAL proposes to learn a reward prediction based task besides a pixel-control based task to speed up learning by improving representation learning. In contrast to on-policy A3C, UNREAL uses an experience replay buffer that is sampled with more priority given to positively rewarded interactions to improve the critic network. Our method, A3C-TP,  differs from UNREAL in several ways: (i) We do not introduce the additional critic improvement step -- to better isolate the gain of our auxiliary task over vanilla A3C. (ii) Even though we also integrate an auxiliary task, we keep the resulting method still on-policy with minimal refinements without an experience replay buffer which might require correction for stale experience data. (iii) UNREAL's reward-prediction requires class balancing of observed rewards in an off-policy fashion depending on the game reward sparsity and distribution whereas \emph{Terminal Prediction} is balanced automatically, it can be applied within on-policy DRL methods, and it generalizes better for episodic tasks independently of the domain-specific reward distribution.

\section{Preliminaries}

We start with the standard reinforcement learning setting of an agent interacting in an environment over a discrete number of steps. At time $t$ the agent in state $s_t$ takes an action $a_t$ and receives a reward $r_t$. The state-value function is the expected return (sum of discounted rewards, ${R_{t:\infty} = \sum_{k=t}^{\infty} \gamma^{k} r_k}$) from state $s$ following a policy $\pi(a|s)$:
$$V^\pi(s)=\mathbb{E}[R_{t:\infty}|s_t=s,\pi],$$
and the action-value function is the expected return following policy $\pi$ after taking action $a$ from state $s$:
$$Q^\pi(s,a)=\mathbb{E}[R_{t:\infty}|s_t=s, a_t=a,\pi].$$

A3C (Asynchronous Advantage Actor-Critic) is an algorithm that employs a \emph{parallelized} asynchronous training scheme (e.g., using multiple CPUs) for efficiency; it is an on-policy RL method that does not need an experience replay buffer. A3C allows multiple workers to simultaneously interact with the environment and compute gradients locally. All the workers pass their computed local gradients to a global network which performs the optimization and synchronizes the updated actor-critic neural network parameters with the workers asynchronously. A3C maintains a parameterized policy (actor) $\pi(a|s;\theta)$ and value function (critic) $V(s; \theta_v)$, which are updated as follows: 
$\triangle \theta = \nabla_\theta \log \pi(a_t|s_t; \theta) A(s_t, a_t; \theta_v) $ and  $\triangle \theta_v = A(s_t, a_t; \theta_v) \nabla_{\theta_v} V(s_t)$ where
$$A(s_t, a_t; \theta_v) = \sum_{k=0}^{n-1} \gamma^kr_{t+k} + \gamma^n V(s_{t+n}) - V (s_t),$$ with $A(s,a)=Q(s,a)-V(s)$ representing the \emph{advantage} function, commonly used to reduce variance.

The policy and the value function are updated after every $t_{max}$ actions or when a terminal state is reached. It is common to use a softmax output layer for the policy head $\pi(a_t|s_t; \theta)$ and one linear output for the value function head $V (s_t; \theta_v)$, with all non-output layers shared, see Figure~\ref{fig:a3c_tp_nn}~(a). 

The loss function for A3C is composed primarily of two terms: policy loss (actor), $\mathcal{L}_{\pi}$, and value loss (critic), $\mathcal{L}_{v}$. An entropy loss for the policy, $H(\pi)$, is also commonly added, which helps to improve exploration by discouraging premature convergence to suboptimal deterministic policies~\cite{mnih2016asynchronous}. Thus, the loss function is given by: $$\mathcal{L}_{\text{A3C}} = \lambda_v  \mathcal{L}_{v} + \lambda_{\pi} \mathcal{L}_{\pi} - \lambda_{H} \mathbb{E}_{s \sim \pi} [H(\pi(s, \cdot, \theta)] $$ with $\lambda_{v}=0.5$, $\lambda_{\pi}=1.0$, and $\lambda_{H}=0.01$, being standard weighting terms on the individual loss components.

\section{Terminal Prediction as an Auxiliary Task}

\begin{figure}
\centering
\includegraphics[scale=0.50]{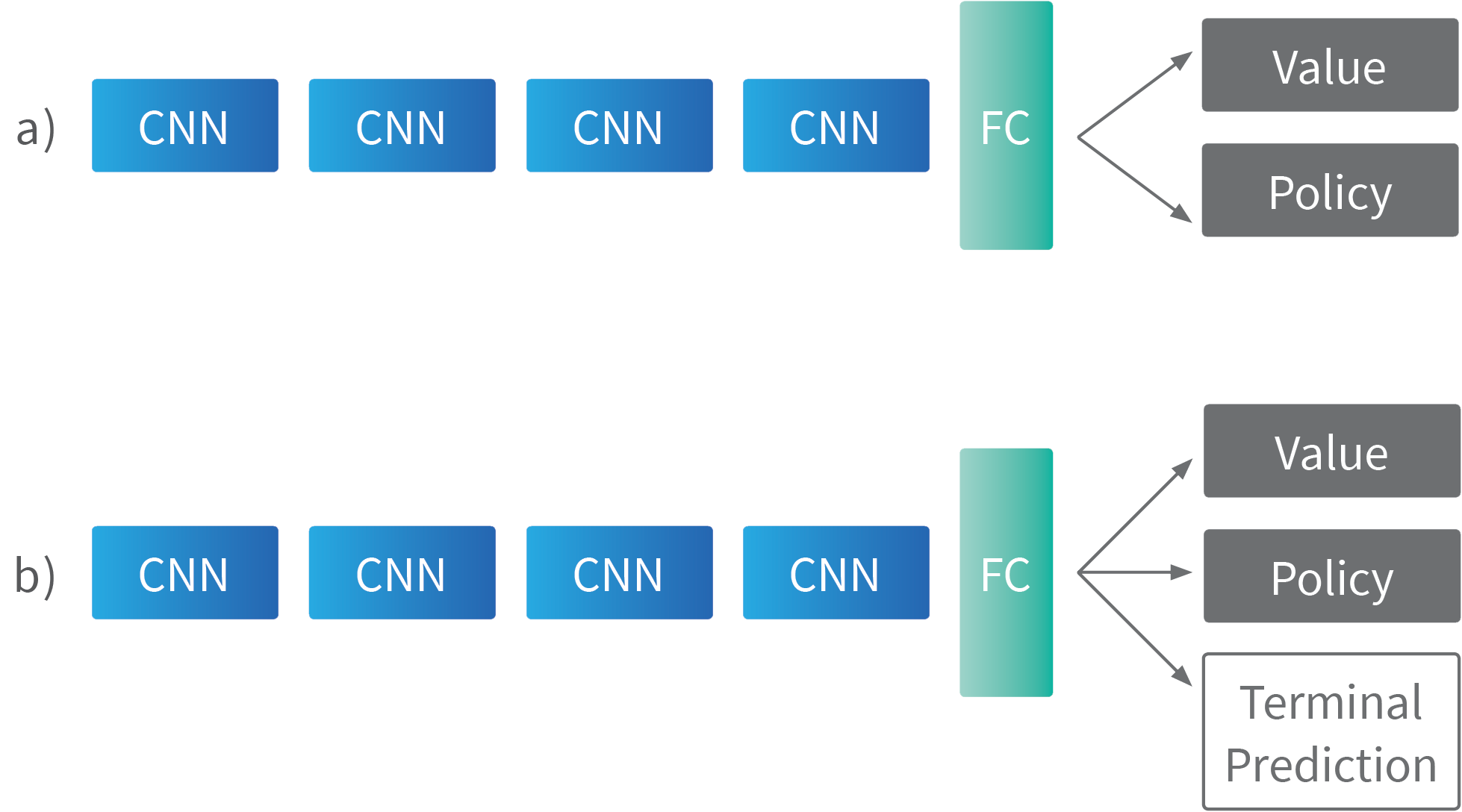}
\caption{CNN represents convolutional neural network layers, and FC represents fully-connected layer. (a) A standard actor-critic neural network architecture for A3C with policy (actor) and value (critic) heads. (b) The neural network architecture of
A3C-TP is identical to that of A3C, except the auxiliary TP head.}
\label{fig:a3c_tp_nn}
\end{figure}

In this section, we describe our main contribution, i.e., the \emph{Terminal Prediction} (TP) auxiliary task. We name our method A3C-TP and present its neural network architecture in Figure~\ref{fig:a3c_tp_nn}~(b). The main motivation is to equip the RL agent with the capability to predict a measure of temporal closeness to terminal states that are more likely to be reached under the agent's current policy.\footnote{Sutton et al.~(\citeyear{sutton2011horde}) mentioned a similar idea when discussing \emph{general value functions}: ``Suppose we are playing a game for which base terminal rewards are +1 for winning and -1 for losing (...) we might pose an independent question about how many more moves the game will last." Interestingly, we only learned about this connection after writing the paper.} We present how simple to compute and integrate this auxiliary task is with deep RL methods, particularly to A3C.

For the proposed auxiliary task of \emph{Terminal Prediction}, i.e., for each observation under the current policy, the agent predicts a temporal closeness value to a terminal state for the current episode under the agent's current policy. The neural network architecture of A3C-TP is identical to that of A3C, fully sharing parameters, except the additional terminal state prediction head.

We compute the loss term for the terminal state prediction head, $\mathcal{L}_{TP}$, by using mean squared error between the predicted value of closeness to a terminal state of any given state (i.e., $y^p$) and the target values approximately computed from completed episodes (i.e., $y$) as follows: $${\mathcal{L}_{TP}= \frac{1}{N} \sum_{i=0}^{N}(y_{i} - y_{i}^p)^2}$$ 
where $N$ represents the average length of previous episode lengths during training. We assume that the target for $i$th state can be approximated with $y_{i} = i/N $ implying $y_{N}=1$ for the actual terminal state and $y_{0}=0$ for the initial state for each episode, and intermediate values are linearly interpolated between $[0,1]$. 

We initially used the actual current episode length for $N$ to compute TP labels. However, this delays access to the labels until the episode is over, similar to observing an external reward signal only when the episode ends. Furthermore, it did not provide significant improvement. To be able to have access to dense signals through our auxiliary task, we decided to use the \emph{running average} of episode lengths computed from the most recent $100$ episodes, to approximate the actual current episode length for $N$. We observed that the running average episode length does not abruptly change, but it can vary in correlation with the learned policy improvements. This approximation not only provides learning performance improvement, but also provides significant memory efficiency for distributed on-policy deep RL as CPU workers do not have to retain the computation graph, i.e. used within deep learning frameworks,  until episode termination to compute terminal prediction loss.

Given $\mathcal{L}_{TP}$, we define the loss for A3C-TP as follows: $$\mathcal{L}_{\text{A3C-TP}}= \mathcal{L}_{\text{A3C}} + \lambda_{TP} \mathcal{L}_{TP}$$ where $\lambda_{TP}$ is a weight term that we set to $0.5$ from experimental results (presented in the next section) to enable the agent to mainly focus on optimizing the policy.

The hypothesis is that the terminal state prediction provides some grounding to the neural network during learning with a denser signal as the agent learns not only to maximize rewards but it can also predict approximately how close it is to the end of episode.

\section{Experiments}

First, we describe our experimental domains, and then we present results comparing our proposed A3C-TP with A3C. 

\subsection{Domains and Setup}

We trained on 6 Atari games (with discrete actions), the BipedalWalker domain (with continuous actions), and Pommerman (with discrete actions and two different opponents). In all cases, we trained both standard A3C and A3C-TP with 3 different random seeds. For Atari games and the BipedalWalker domain, we employed only 8 CPU workers. For the Pommerman domain, we increased to 24 CPU workers as Pommerman is very challenging for model-free RL.

\subsubsection{Atari Games and BipedalWalker}

We conducted experiments using OpenAI Gym~\cite{brockman2016openai} using sticky actions and stochastic frame-skipping. We did not perform any reward scaling across the games~\cite{chrabaszcz2018back} so that agents can distinguish the reward magnitudes rather than just being rewarded with $+1$ or $-1$.

The Atari benchmark provides several games with diverse challenges that render different methods more successful in different subset of games,\footnote{Emekligil et al.~(\citeyear{emekligils}) recently analyzed which factors contribute to the Deep RL methods' success within Atari games.} e.g., the game Frostbite is mastered by DQN, but A3C performs poorly on it. Thus, we aimed to select a small but diverse set of games with varying difficulties for computational cost reasons. The selected games are: Pong, Breakout, CrazyClimber, Q*bert, BeamRider, and SpaceInvaders. The first 3 games are simpler games where dense reward signals are ubiquitous and reactive strategies learned by the RL method provide high rewards to progress the policy learning; Q*bert, Seaquest, and SpaceInvaders are relatively harder as longer term non-reactive strategies are needed~\cite{mnih2015human,mnih2016asynchronous,chrabaszcz2018back}. 

We also experimented with the BipedalWalker (available through OpenAI Gym), which takes only a few hours to train on a standard laptop. Main difference of this domain is that it is a continuous action-space one.

\subsubsection{Pommerman}

\begin{figure}
\centering
\includegraphics[scale=0.36]{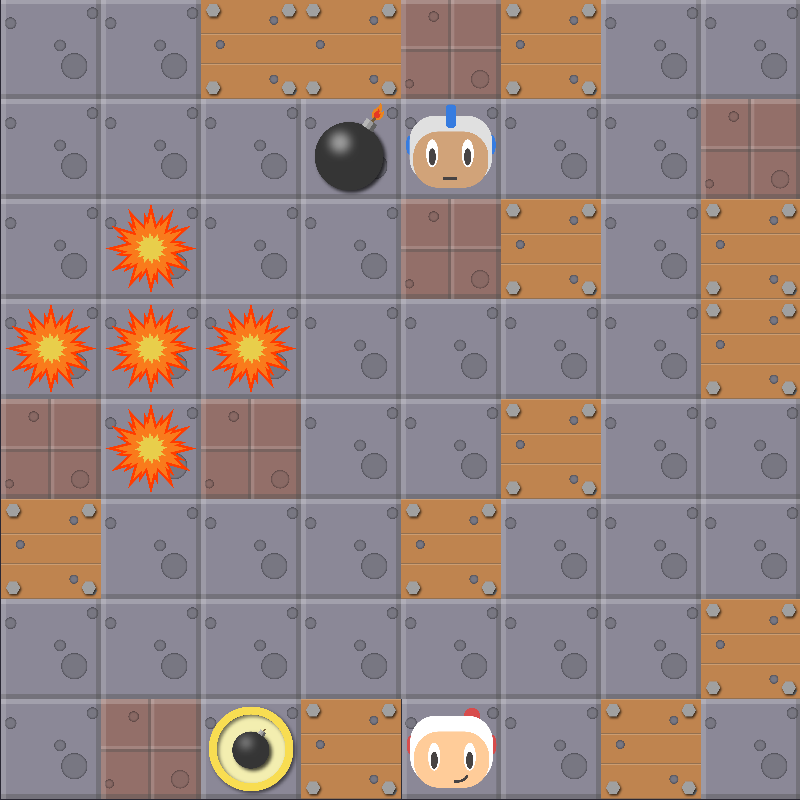}
\caption{An example of the $8 \times 8$ Pommerman board, randomly generated by the simulator. Agents' initial positions are randomly selected among four corners at each episode.}
\label{fig:pom8x8}
\end{figure}

The Pommerman environment~\cite{resnick2018pommerman} is based off of the classic console game Bomberman, played with four agents on an 11x11 board. Our experiments use a mini version of the simulator with only two agents on an 8x8 board (see Figure~\ref{fig:pom8x8}). Each agent can execute one of 6 actions at every timestep: move in any of cardinal directions, stay put, or place a bomb. Each cell on the board can be a passage, a rigid wall, or wood. The maps are generated randomly, albeit there is always a guaranteed path between any two agents. Whenever an agent places a bomb it explodes after 10 timesteps, producing flames that have a lifetime of 2 timesteps. Flames destroy wood and kill any agents within their blast radius. When wood is destroyed either a passage or a power-up is revealed.  Power-ups can be of three types: increase the blast radius of bombs, increase the number of bombs the agent can place, or give the ability to kick bombs. A single game is finished when an agent dies or when reaching 800 timesteps.

Pommerman is a very challenging benchmark for RL methods. The first challenge is that of sparse and delayed rewards: the game can last up to 800 timesteps and the environment only provides terminal reward per episode. A second issue is the complex environment since tile locations and agents' initial locations are randomized at the beginning of every game (episode). The game board changes within each episode too due to the (dis)appearance of the wood, power-ups, flames, and bombs. The last complication is the multiagent component since the agent needs to best respond to any type of opponent whose behaviour could change based on collected power-ups.

Also, the rewards in Pommerman game can be deceiving. We consider false positive episodes when our agent gets a positive reward because the opponent commits suicide (not due to our agent's combat skill), and false negative episodes when our agent gets a negative reward due to its own suicide (rather than getting killed by an enemy bomb). False negative episodes are a major bottleneck for learning reasonable behaviours with pure-exploration within the RL. Besides, false positive episodes also can reward agents for arbitrary passive survival policies such as camping rather than engaging actively with opponents.

For these reasons, generally a local optimum is commonly learned, i.e., not placing bombs~\cite{resnick2018pommerman}.

We considered two types of opponents in this domain:
\begin{itemize}
    \item \emph{Static} opponents: the opponent waits in the initial position and always executes the `stay put' action. This opponent is the simplest possible opponent  as it provides a more stationary environment.
    
    \item \emph{Rule-based} opponents: the baseline agent that comes within the simulator. It collects power-ups and places bombs when it is near an opponent. It is skilled in avoiding blasts from bombs. It uses Dijkstra's algorithm on each time-step, resulting in longer training times. Its behaviour is highly stochastic based on the power-ups it has collected, e.g., if it collected certain power-ups, it can place many bombs triggering chain explosions (bombs explode earlier due to being on a flame zone created by another exploding bomb).
\end{itemize}

\subsection{Implementation Details} 
\textbf{NN Architecture and Hyperparameters:} For Atari and BipedalWalker domains, our NN architecture uses 4 convolutional layers where the first two have 32 filters with a kernel size of $5 \times 5$, and the remaining ones have 64 filters with a stride and padding of 1 with a filter size of $4 \times 4$ and $3 \times 3$. This is followed with an LSTM layer with 128 units leading to standard actor-critic heads along with our \emph{Terminal Prediction head}. 
For Pommerman, our NN architecture similarly uses 4 convolutional layers, each of which has 32 filters and $3 \times 3$ filters with a stride and padding size of 1, followed with a fully connected layer with 128 hidden units which leads to the three output heads. For both domains, we employed the Adam optimizer with a learning rate of $0.0001$. Neural network architectures are not tuned. 

\textbf{NN State Representation:} For Atari games, the visual input is fed to the NN as an input whereas in Pommerman, similar to ~\cite{resnick2018pommerman}, we maintain 22 feature maps that are constructed from the agent observation. These feature channels maintain location of walls, wood, power-ups, agents, bombs, and flames. Agents have different properties such as bomb kick, bomb blast radius, and number of bombs. We maintain 3 feature maps for these abilities per agent. We also maintain a feature map for the remaining lifetime of flames.

\subsection{Results}

In this section, we first present a sensitivity analysis on the \emph{Terminal Prediction} loss weight term. Then, we show training results for both Atari and Pommerman domains to ablate the contribution of our proposed method, i.e., A3C-TP, against the standard A3C.

\begin{figure}
    \centering
    \subfloat[Pong]{{\includegraphics[scale=0.27]{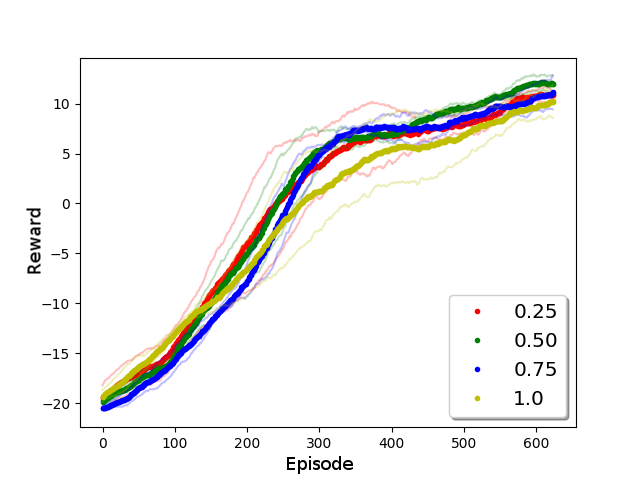} }} 
    \subfloat[Breakout]{{\includegraphics[scale=0.27]{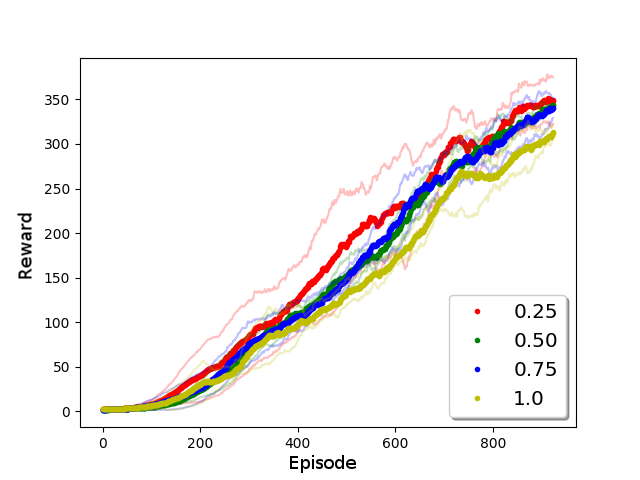}}}%
    \caption{\textbf{Sensitivity Analysis on $\lambda_{TP}$:} Moving average over 100 games of the rewards (horizontal lines depict individual episodic rewards) is shown. Each training curve depicts the average and standard deviation of 3 experiments with different random seeds. Both games are trained for a day using 8 CPU cores.}%
    \label{fig:sensitivity_tp}%
\end{figure}

\paragraph{Sensitivity Analysis on $\lambda_{TP}$} We conducted preliminary experiments on two (fast to train) Atari games, i.e., Pong and Breakout. We tested learning performance on 4 values $\{0.25, 0.5, 0.75, 1\}$ for the \emph{Terminal Prediction} weight. Each curve is obtained from training with 3 random seeds. As Figure~\ref{fig:sensitivity_tp} shows, the learning performance gets worse when $\lambda_{TP}=1$ as the agent puts too much emphasis on learning the auxiliary task compared to the main policy learning task. On the other hand, when $\lambda_{TP}=0.25$ the variance is higher for both games. Given this preliminary analysis where intermediate values of $\{0.5,0.75\}$ perform more reliably, we chose to employ $\lambda_{TP}=0.5$ for all the experiments. 

\subsubsection{Atari Games}

\begin{figure*}
    \centering
    \subfloat[Q*bert]{{\includegraphics[scale=0.21]{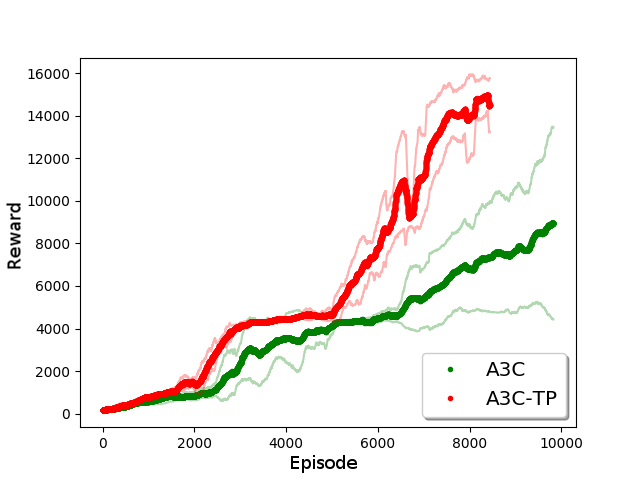} }}%
    \subfloat[CrazyClimber]{{\includegraphics[scale=0.21]{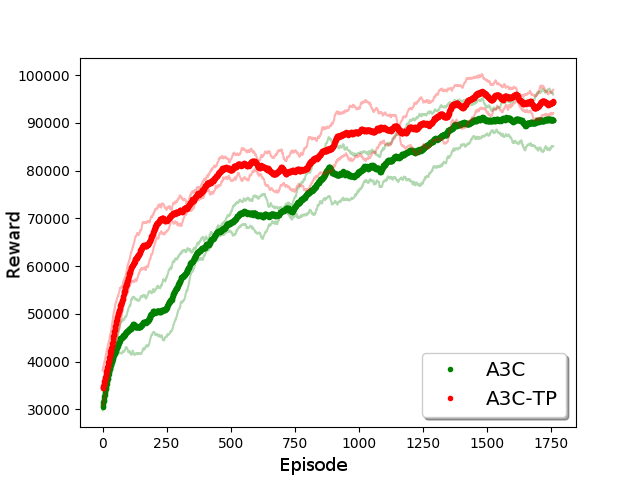}}}%
    \subfloat[SpaceInvaders]{{\includegraphics[scale=0.21]{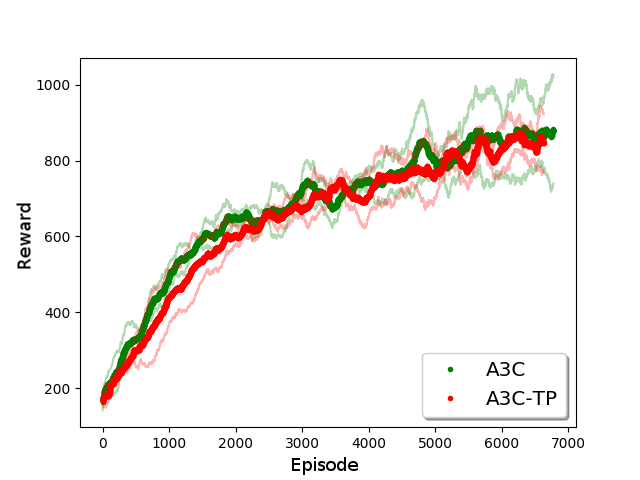} }}%
    \subfloat[BeamRider]{{\includegraphics[scale=0.21]{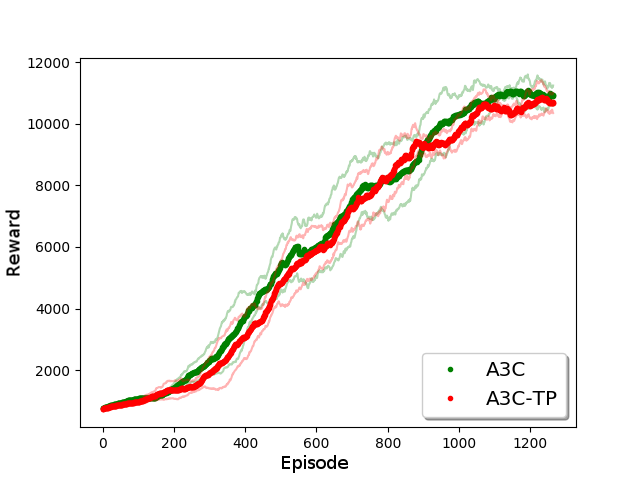} }} 
    \subfloat[BipedalWalker]{{\includegraphics[scale=0.21]{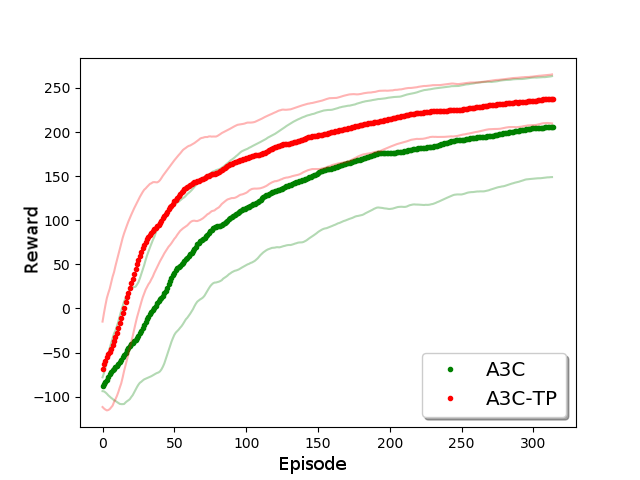} }} 
    \caption{ Moving average over 50 games of the rewards (horizontal lines depict individual episodic rewards) is shown. Atari games are trained for 3 days, BipedalWalker is trained for 5 hours, all using 8 CPU workers with 3 random seeds. Our method, A3C-TP, performed no worse than standard A3C in any tested games, and it outperformed A3C in Q*bert and CrazyClimber games, and BipedalWalker domain.}%
    \label{fig:atari_results}%
\end{figure*}

We present comparisons of our proposed method, A3C-TP against standard A3C method in Figure~\ref{fig:atari_results}. Our method significantly improves upon standard A3C in Q*bert and CrazyClimber games while performing similarly in the rest of the tested games. As we employed relatively easy games Pong and Breakout games for sensitivity analysis for A3C-TP, we omitted them from Figure~\ref{fig:atari_results}. Our hypothesis is that A3C-TP does not hurt the policy learning performance, however it can provide improvement in some benchmarks.

\subsubsection{BipedalWalker}

We also benchmark on a continuous action-space domain, BipedalWalker. The experimental results in Figure~\ref{fig:atari_results} (e) show that A3C-TP outperforms A3C, and it has relatively lower variance.

\subsubsection{Pommerman}

\begin{figure}
    \centering
    \subfloat[Learning against a \emph{Static} opponent.]{{\includegraphics[scale=0.32]{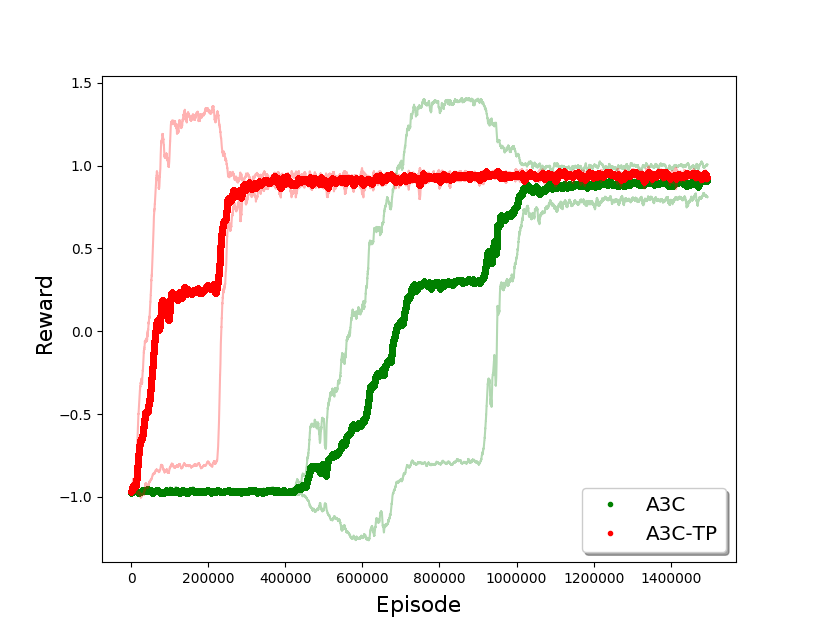} }} \\
    \subfloat[Learning against a \emph{Rule-based} opponent]{{\includegraphics[scale=0.32]{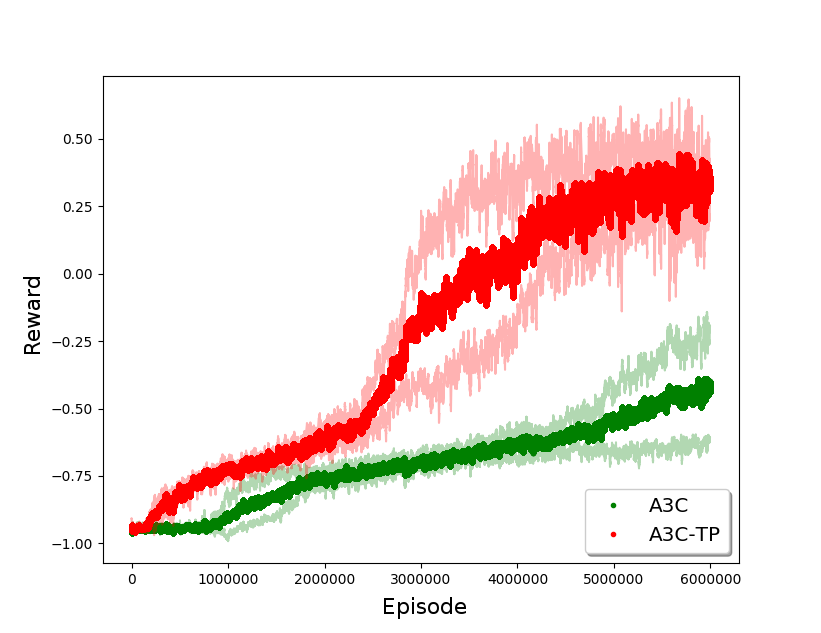} }}%
    \caption{\textbf{Pommerman Domain Results:} Moving average over 5k games of the rewards (horizontal lines depict individual episodic rewards) is shown. Each training curve depicts average and standard deviation of 3 experiments with different random seeds. Our method, A3C-TP, outperforms the standard A3C in terms of both learning faster and converging to a better policy in learning against both \emph{Static} and \emph{Rule-based} opponents. The training times was 6 hours for (a) and 3 days for (b) both using 24 CPU cores.}%
    \label{fig:a3c_static_simple}%
\end{figure}

We present comparison results based on the training performance in terms of converged policies and time-efficiency against \emph{Static} and \emph{Rule-based} opponents.

The training results for A3C and our proposed method, A3C-TP, against a \emph{Static} opponent are shown in Figure~\ref{fig:a3c_static_simple}~(a). Our method both converges much faster compared to the standard A3C. As \emph{Static} opponents do not commit suicide, there are no false positives in observed positive rewards. However there are still false negatives due to our agent's possible suicides during training that negatively reinforces the essentially needed bombing skill. In this scenario, A3C takes almost one million episodes to converge because the agent needs to learn to execute a series of actions (without killing itself): get close to the enemy, place a bomb near the enemy and stay alive until the bomb explodes in 10 timesteps, eliminating the enemy. Only after successful execution of these actions, the agent obtains a non-zero external reward signal. One reason we present results and analysis against \emph{Static} opponents is to convey how challenging the Pommerman domain is for purely model-free RL.

Against the \emph{Rule-based} opponent our method learns faster, and it finds a better policy in terms of average rewards, see Figure~\ref{fig:a3c_static_simple}~(b). Training against \emph{Rule-based} opponents takes much longer, possibly due to episodes with false positive rewards.

\section{Discussion and Future Work}

Atari games have been heavily used by RL researchers, but Pommerman was recently released and there are many open problems. Some recent works have used Pommerman as a test-bed, e.g. Zhou et al.~(\citeyear{zhou2018hybrid}) proposed a hybrid method combining rule-based heuristics with depth-limited search. Resnick et al.~(\citeyear{resnick2018backplay}) proposed a framework that uses a single demonstration to generate a training curriculum for sparse reward RL problems, including Pommerman, assuming episodes can be started from arbitrary states. In this work, we use this highly challenging domain and show that our proposed method provides significant improvement for sample efficiency.

We want to note that there are already a variety of auxiliary tasks proposed in the literature, and they are in general heuristics proposed as additional self-supervised learning targets to help RL agents to improve its representation learning with more contextual information.
A recent work shows that using some auxiliary tasks can result in worse performance for the main task~\cite{du2018adapting}. Thus, depending on the domain properties, different auxiliary tasks are likely to have varying performances. However, we note that more theoretical understanding of auxiliary tasks and how to devise them automatically is still an open research avenue~\cite{bellemare2019geometric,sutton2018reinforcement}. 

As future research, we think TP could be employed for safe exploration and within the context of learning from demonstrations. TP values could be used to bias the exploration strategy within the context of safe RL~\cite{garcia2015comprehensive}. For example, the RL agent could combine state-value function estimates with TP estimate: it can trigger exploitation when both state-value and TP estimates are high (e.g., a win case) whereas exploration could be triggered when state-value estimate is low and TP estimate is high, implying for example, a dead-end for the agent.

Another avenue we want to investigate is to use TP values for better integration of demonstrators and model-free RL within the context of imitation learning. For example, a (search-based, simulated) demonstrator can be asked for action-guidance when the TP value is high and state-value estimate is low, i.e., implying the episode is predicted to be close to a terminal state against the favor of the agent. TP estimates could also be used to determine a reward discounting schedule or the effective horizon length to consider future rewards~\cite{fedus2019hyperbolic}. One main limitation of \emph{Terminal Prediction} auxiliary task is its  applicability to only episodic tasks (in contrast to other existing auxiliary tasks).

\section{Conclusions}

Deep reinforcement learning has achieved great successes in recent years with the help of novel methods and higher compute power. However, DRL is often sample inefficient. One of the active research area to address this challenge is to use auxiliary tasks. Along this line of research, in this paper, we propose a novel self-supervised auxiliary task, i.e., \emph{Terminal Prediction (TP)}, estimating the temporal closeness to terminal states for episodic tasks, which can be easily integrated to existing DRL methods to improve the learning efficiency. We experimented integrating this task with A3C proposing A3C-TP which improves the policy learning performance across different domains (e.g., Atari, BipedalWalker, and Pommerman) with diverse dynamics. Our TP task is general as can be easily integrated with other DRL methods and with any episodic domain.

\bibliographystyle{aaai}
\bibliography{ref}

\begin{thebibliography}{}

\bibitem[\protect\citeauthoryear{Arulkumaran \bgroup et al\mbox.\egroup
  }{2017}]{arulkumaran2017deep}
Arulkumaran, K.; Deisenroth, M.~P.; Brundage, M.; and Bharath, A.~A.
\newblock 2017.
\newblock Deep reinforcement learning: A brief survey.
\newblock {\em IEEE Signal Processing Magazine} 34(6):26--38.

\bibitem[\protect\citeauthoryear{Bellemare \bgroup et al\mbox.\egroup
  }{2019}]{bellemare2019geometric}
Bellemare, M.~G.; Dabney, W.; Dadashi, R.; Taiga, A.~A.; Castro, P.~S.; Roux,
  N.~L.; Schuurmans, D.; Lattimore, T.; and Lyle, C.
\newblock 2019.
\newblock A geometric perspective on optimal representations for reinforcement
  learning.
\newblock {\em arXiv preprint arXiv:1901.11530}.

\bibitem[\protect\citeauthoryear{Brockman \bgroup et al\mbox.\egroup
  }{2016}]{brockman2016openai}
Brockman, G.; Cheung, V.; Pettersson, L.; Schneider, J.; Schulman, J.; Tang,
  J.; and Zaremba, W.
\newblock 2016.
\newblock Open{AI} gym.
\newblock {\em arXiv preprint arXiv:1606.01540}.

\bibitem[\protect\citeauthoryear{Chrabaszcz, Loshchilov, and
  Hutter}{2018}]{chrabaszcz2018back}
Chrabaszcz, P.; Loshchilov, I.; and Hutter, F.
\newblock 2018.
\newblock Back to basics: Benchmarking canonical evolution strategies for
  playing atari.
\newblock {\em arXiv preprint arXiv:1802.08842}.

\bibitem[\protect\citeauthoryear{Du \bgroup et al\mbox.\egroup
  }{2018}]{du2018adapting}
Du, Y.; Czarnecki, W.~M.; Jayakumar, S.~M.; Pascanu, R.; and Lakshminarayanan,
  B.
\newblock 2018.
\newblock Adapting auxiliary losses using gradient similarity.
\newblock {\em arXiv preprint arXiv:1812.02224}.

\bibitem[\protect\citeauthoryear{Emekligil and Alpayd{\i}n}{2018}]{emekligils}
Emekligil, E., and Alpayd{\i}n, E.
\newblock 2018.
\newblock What’s in a game? the effect of game complexity on deep
  reinforcement learning.
\newblock In {\em Workshop on Computer Games},  147--163.

\bibitem[\protect\citeauthoryear{Fedus \bgroup et al\mbox.\egroup
  }{2019}]{fedus2019hyperbolic}
Fedus, W.; Gelada, C.; Bengio, Y.; Bellemare, M.~G.; and Larochelle, H.
\newblock 2019.
\newblock Hyperbolic discounting and learning over multiple horizons.
\newblock {\em arXiv preprint arXiv:1902.06865}.

\bibitem[\protect\citeauthoryear{Fran{\c{c}}ois-Lavet \bgroup et
  al\mbox.\egroup }{2018}]{franccois2018introduction}
Fran{\c{c}}ois-Lavet, V.; Henderson, P.; Islam, R.; Bellemare, M.~G.; Pineau,
  J.; et~al.
\newblock 2018.
\newblock An introduction to deep reinforcement learning.
\newblock {\em Foundations and Trends in Machine Learning} 11(3-4):219--354.

\bibitem[\protect\citeauthoryear{Garc{\i}a and
  Fern{\'a}ndez}{2015}]{garcia2015comprehensive}
Garc{\i}a, J., and Fern{\'a}ndez, F.
\newblock 2015.
\newblock A comprehensive survey on safe reinforcement learning.
\newblock {\em Journal of Machine Learning Research} 16(1):1437--1480.

\bibitem[\protect\citeauthoryear{Hernandez-Leal, Kartal, and
  Taylor}{2018}]{hernandez2018multiagent}
Hernandez-Leal, P.; Kartal, B.; and Taylor, M.~E.
\newblock 2018.
\newblock {Is multiagent deep reinforcement learning the answer or the
  question? A brief survey}.
\newblock {\em arXiv preprint arXiv:1810.05587}.

\bibitem[\protect\citeauthoryear{Jaderberg \bgroup et al\mbox.\egroup
  }{2017}]{jaderberg2016reinforcement}
Jaderberg, M.; Mnih, V.; Czarnecki, W.~M.; Schaul, T.; Leibo, J.~Z.; Silver,
  D.; and Kavukcuoglu, K.
\newblock 2017.
\newblock Reinforcement learning with unsupervised auxiliary tasks.
\newblock In {\em International Conference on Learning Representations}.

\bibitem[\protect\citeauthoryear{Konda and Tsitsiklis}{2000}]{konda2000actor}
Konda, V.~R., and Tsitsiklis, J.~N.
\newblock 2000.
\newblock Actor-critic algorithms.
\newblock In {\em Advances in neural information processing systems},
  1008--1014.

\bibitem[\protect\citeauthoryear{Lample and Chaplot}{2017}]{lample2017playing}
Lample, G., and Chaplot, D.~S.
\newblock 2017.
\newblock {Playing FPS Games with Deep Reinforcement Learning}.
\newblock In {\em AAAI},  2140--2146.

\bibitem[\protect\citeauthoryear{LeCun, Bengio, and
  Hinton}{2015}]{lecun2015deep}
LeCun, Y.; Bengio, Y.; and Hinton, G.
\newblock 2015.
\newblock Deep learning.
\newblock {\em Nature} 521(7553):436.

\bibitem[\protect\citeauthoryear{Leibfried and
  Vrancx}{2018}]{leibfried2018modelbased}
Leibfried, F., and Vrancx, P.
\newblock 2018.
\newblock Model-based regularization for deep reinforcement learning with
  transcoder networks.
\newblock {\em arXiv preprint 1809.01906}.

\bibitem[\protect\citeauthoryear{Leibfried, Kushman, and
  Hofmann}{2016}]{leibfried2016deep}
Leibfried, F.; Kushman, N.; and Hofmann, K.
\newblock 2016.
\newblock A deep learning approach for joint video frame and reward prediction
  in atari games.
\newblock In {\em ICML 2017 Workshop on Principled Approaches to Deep
  Learning}.

\bibitem[\protect\citeauthoryear{Mirowski \bgroup et al\mbox.\egroup
  }{2016}]{mirowski2016learning}
Mirowski, P.; Pascanu, R.; Viola, F.; Soyer, H.; Ballard, A.~J.; Banino, A.;
  Denil, M.; Goroshin, R.; Sifre, L.; Kavukcuoglu, K.; et~al.
\newblock 2016.
\newblock Learning to navigate in complex environments.
\newblock In {\em ICLR}.

\bibitem[\protect\citeauthoryear{Mnih \bgroup et al\mbox.\egroup
  }{2015}]{mnih2015human}
Mnih, V.; Kavukcuoglu, K.; Silver, D.; Rusu, A.~A.; Veness, J.; Bellemare,
  M.~G.; Graves, A.; Riedmiller, M.; Fidjeland, A.~K.; Ostrovski, G.; et~al.
\newblock 2015.
\newblock Human-level control through deep reinforcement learning.
\newblock {\em Nature} 518(7540):529.

\bibitem[\protect\citeauthoryear{Mnih \bgroup et al\mbox.\egroup
  }{2016}]{mnih2016asynchronous}
Mnih, V.; Badia, A.~P.; Mirza, M.; Graves, A.; Lillicrap, T.; Harley, T.;
  Silver, D.; and Kavukcuoglu, K.
\newblock 2016.
\newblock Asynchronous methods for deep reinforcement learning.
\newblock In {\em International conference on machine learning},  1928--1937.

\bibitem[\protect\citeauthoryear{Oh \bgroup et al\mbox.\egroup
  }{2015}]{oh2015action}
Oh, J.; Guo, X.; Lee, H.; Lewis, R.~L.; and Singh, S.
\newblock 2015.
\newblock Action-conditional video prediction using deep networks in atari
  games.
\newblock In {\em NIPS},  2863--2871.

\bibitem[\protect\citeauthoryear{Resnick \bgroup et al\mbox.\egroup
  }{2018}]{resnick2018pommerman}
Resnick, C.; Eldridge, W.; Ha, D.; Britz, D.; Foerster, J.; Togelius, J.; Cho,
  K.; and Bruna, J.
\newblock 2018.
\newblock Pommerman: A multi-agent playground.
\newblock {\em AIIDE Multi-Agent Workshop}.

\bibitem[\protect\citeauthoryear{Resnick \bgroup et al\mbox.\egroup
  }{2019}]{resnick2018backplay}
Resnick, C.; Raileanu, R.; Kapoor, S.; Peysakhovich, A.; Cho, K.; and Bruna, J.
\newblock 2019.
\newblock Backplay:" man muss immer umkehren".
\newblock {\em AAAI-19 Workshop on RL in Games}.

\bibitem[\protect\citeauthoryear{Sharifi, Zhao, and
  Szafron}{2010}]{sharifi2010learning}
Sharifi, A.; Zhao, R.; and Szafron, D.~A.
\newblock 2010.
\newblock Learning companion behaviors using reinforcement learning in games.
\newblock In {\em Sixth Artificial Intelligence and Interactive Digital
  Entertainment Conference}.

\bibitem[\protect\citeauthoryear{Shelhamer \bgroup et al\mbox.\egroup
  }{2016}]{shelhamer2016loss}
Shelhamer, E.; Mahmoudieh, P.; Argus, M.; and Darrell, T.
\newblock 2016.
\newblock Loss is its own reward: Self-supervision for reinforcement learning.
\newblock {\em arXiv preprint arXiv:1612.07307}.

\bibitem[\protect\citeauthoryear{Silver \bgroup et al\mbox.\egroup
  }{2017}]{silver2017mastering}
Silver, D.; Schrittwieser, J.; Simonyan, K.; Antonoglou, I.; Huang, A.; Guez,
  A.; Hubert, T.; Baker, L.; Lai, M.; Bolton, A.; et~al.
\newblock 2017.
\newblock Mastering the game of go without human knowledge.
\newblock {\em Nature} 550(7676):354.

\bibitem[\protect\citeauthoryear{Suddarth and
  Kergosien}{1990}]{suddarth1990rule}
Suddarth, S.~C., and Kergosien, Y.
\newblock 1990.
\newblock Rule-injection hints as a means of improving network performance and
  learning time.
\newblock In {\em Neural Networks}. Springer.
\newblock  120--129.

\bibitem[\protect\citeauthoryear{Sutton and
  Barto}{2018}]{sutton2018reinforcement}
Sutton, R.~S., and Barto, A.~G.
\newblock 2018.
\newblock {\em Reinforcement learning: An introduction}.

\bibitem[\protect\citeauthoryear{Sutton \bgroup et al\mbox.\egroup
  }{2011}]{sutton2011horde}
Sutton, R.~S.; Modayil, J.; Delp, M.; Degris, T.; Pilarski, P.~M.; White, A.;
  and Precup, D.
\newblock 2011.
\newblock Horde: A scalable real-time architecture for learning knowledge from
  unsupervised sensorimotor interaction.
\newblock In {\em AAMAS}.

\bibitem[\protect\citeauthoryear{Wang \bgroup et al\mbox.\egroup
  }{2017}]{wang2017interactive}
Wang, P.; Rowe, J.~P.; Min, W.; Mott, B.~W.; and Lester, J.~C.
\newblock 2017.
\newblock Interactive narrative personalization with deep reinforcement
  learning.
\newblock In {\em IJCAI},  3852--3858.

\bibitem[\protect\citeauthoryear{Watkins and Dayan}{1992}]{watkins1992q}
Watkins, C.~J., and Dayan, P.
\newblock 1992.
\newblock Q-learning.
\newblock {\em Machine learning} 8(3-4):279--292.

\bibitem[\protect\citeauthoryear{Williams}{1992}]{williams1992simple}
Williams, R.~J.
\newblock 1992.
\newblock Simple statistical gradient-following algorithms for connectionist
  reinforcement learning.
\newblock {\em Machine learning} 8(3-4):229--256.

\bibitem[\protect\citeauthoryear{Yu}{2018}]{yu2018towards}
Yu, Y.
\newblock 2018.
\newblock Towards sample efficient reinforcement learning.
\newblock In {\em IJCAI},  5739--5743.

\bibitem[\protect\citeauthoryear{Zhou \bgroup et al\mbox.\egroup
  }{2018}]{zhou2018hybrid}
Zhou, H.; Gong, Y.; Mugrai, L.; Khalifa, A.; Nealen, A.; and Togelius, J.
\newblock 2018.
\newblock A hybrid search agent in pommerman.
\newblock In {\em Proceedings of the 13th International Conference on the
  Foundations of Digital Games}.

\end{thebibliography}

\end{document}